\documentclass[10pt,twocolumn,letterpaper]{article}

\usepackage[pagenumbers]{cvpr} 

\usepackage{pifont}
\newcommand{\cmark}{\ding{51}} 
\newcommand{\xmark}{\ding{55}} 
\usepackage{multirow}
\usepackage{float}
\usepackage[table]{xcolor}

\definecolor{cvprblue}{rgb}{0.2,0.2,0.8} 
\usepackage[pagebackref,breaklinks,colorlinks,allcolors=cvprblue]{hyperref}

\def\paperID{*****} 
\def\confName{CVPR}
\def\confYear{2026}

\title{CoDeFont: One-Shot Font Generation via Multi-Scale Factor-Specific Representation Learning}

\author{
Rejoy Chakraborty$^{1}$ \qquad
Prasun Roy$^{1}$ \qquad
Saumik Bhattacharya$^{2}$ \qquad
Umapada Pal$^{1}$ \vspace{0.5em}\\
$^{1}$Indian Statistical Institute Kolkata \qquad $^{2}$Indian Institute of Technology Kharagpur \vspace{0.5em}\\
{\tt\small\textbf{\url{https://rejoycs.github.io/codefont}}}
}

\begin{document}


\twocolumn[{%
\renewcommand\twocolumn[1][]{#1}%
\maketitle
}]

\begin{abstract}
One-shot Font Generation (OFG) aims to synthesize a target glyph by transferring the stylistic characteristics of a single reference glyph while preserving the structure of the target character. However, accurately capturing font style from a single reference image remains challenging due to the limited stylistic information available in a single glyph, and this challenge is further amplified for complex scripts. We propose CoDeFont, a script-agnostic OFG framework that addresses this challenge through multi-level style-content factorization and progressive style-content reconciliation. CoDeFont independently aggregates complementary style and structural information across multiple spatial levels to obtain factor-specific representations, while a contrastive objective explicitly encourages separation between style and content embeddings. During reconstruction, the disentangled representations are progressively injected through multi-level style conditioning, enabling stylistic information to be incorporated at different decoding resolutions while preserving the underlying glyph structure. We evaluate CoDeFont on structurally distinct Bengali and Latin scripts across seen and unseen font styles. Extensive quantitative and qualitative analyses demonstrate consistent improvements over existing state-of-the-art methods, highlighting the effectiveness of the proposed framework for OFG across diverse writing systems.
\end{abstract}

\section{Introduction}
\label{sec:intro}

In the domain of Image-to-Image (I2I) translation, Few-shot Font Generation (FFG) is an active research problem that aims to synthesize glyphs in a target font style using few reference glyphs. One-shot Font Generation (OFG) is a subdomain of FFG where we fix the set of style references to one. TrueType Font (TTF) files are widely used for digital font rendering, as they encode the geometric structure and stylistic characteristics of a font style. Given a small set of reference glyphs, estimating the underlying TTF is highly challenging because the available references do not provide sufficient information to infer all font-specific properties and glyph outlines. To overcome the issue, FFG methods focus on directly generating the target glyph image.

In the early stages of FFG methods, researchers primarily adopted image-to-image (I2I) style transfer mechanisms to synthesize glyphs in a target font style~\cite{isola2017image, roy2020stefann}. While these approaches demonstrated the feasibility of font style transfer, they often struggled to preserve fine-grained stylistic details, spatial relationships, and structural consistency across generated glyphs. To address these limitations, structure-aware methods~\cite{xie2021dg} were proposed, where the encoder learns spatial correspondences between reference and target glyphs. Recent studies~\cite{zeng2025few} use stroke decomposition to represent glyphs with predefined components, but such representations are often script-dependent. Alternatively, other approaches~\cite{xie2021dg, park2021multiple} disentangle style and structural information by learning separate latent representations.

Despite recent progress, existing approaches largely treat style-content disentanglement, multi-scale representation, and style injection as separate components, lacking a unified mechanism for multi-level factorization and progressive fusion during reconstruction. Further, there are limited methods~\cite{hossain2025bengalidiff} available focusing on structurally complex scripts such as Indic scripts, which feature intricate strokes and multi-zone layouts. To address these, we propose \textbf{CoDeFont}\footnote{Under consideration at Pattern Recognition Letters} (\textbf{Co}ntrastively \textbf{D}is\textbf{e}ntangled \textbf{Font}), a script-agnostic OFG framework that synthesizes high-fidelity glyphs from a single style reference image. Unlike existing approaches that primarily aggregate or propagate representations through a shared reconstruction pathway, CoDeFont explicitly factorizes style using \textbf{SQAH} module and content across multiple representation levels using \textbf{CPH} module, and progressively fuses the disentangled style information with content during decoding through \textbf{MASFM} module. This design enables effective preservation of font-specific stylistic characteristics while maintaining the structural integrity of target glyphs.

\subsection*{Contributions:}

\begin{enumerate}
\item We propose \textbf{CoDeFont}, a one-shot font generation framework that formulates font synthesis as multi-level style-content factorization, aggregating rich representations and progressively reconciling them during glyph reconstruction without relying on script-specific information.

\item We introduce a factor-specific representation learning strategy that combines contrastive and auxiliary classification objectives to disentangle style and content embeddings, yielding strong factor-specificity validated through embedding-space analysis.

\item Extensive qualitative and quantitative experiments demonstrate that the proposed method consistently outperforms existing state-of-the-art (SOTA) font generation approaches on Bengali and Latin scripts, demonstrating its effectiveness across scripts with distinct structural characteristics.
\end{enumerate}

\noindent
The remainder of this paper is organized as follows. Section~\ref{sec:related} reviews the relevant literature, Section~\ref{sec:method} presents the proposed methodology, Section~\ref{sec:results} reports the comparative results and analysis, and Section~\ref{sec:conclusion} concludes the paper with a summary of the key findings and future directions.

\section{Related Works} \label{sec:related}

\subsection{Image-to-Image Translation}
Image-to-Image (I2I) translation methods primarily aim to transform content from the source domain to the target domain.  With the emergence of GAN~\cite{goodfellow2014generative}, several I2I methods~\cite{yi2017dualgan, choi2018stargan} based on GANs have been proposed. Pix2Pix~\cite{isola2017image} introduced a conditional GAN pipeline using paired image data. Later, DualGAN~\cite{yi2017dualgan} proposed an unsupervised I2I pipeline using unlabeled data. CycleGAN~\cite{zhu2017unpaired} enables unsupervised image-to-image translation through cycle-consistency learning, whereas FUNIT~\cite{liu2019few} synthesizes target-domain images by conditioning content features on style embeddings using AdaIN~\cite{huang2017arbitrary}. Later, a few methods like~\cite{ li2023bbdm, tumanyan2023plug} incorporated a diffusion-based method to handle complex fonts.

\subsection{Few-shot Font Generation}
Early font generation methods~\cite{zhou2011easy, campbell2014learning} primarily relied on handcrafted structural representations, such as strokes and radicals. Although effective for regular typefaces, these approaches struggle to model complex artistic styles. With the advent of deep generative models~\cite{yang2017awesome}, substantial progress has been achieved in font synthesis. STEFANN~\cite{roy2020stefann} pioneered scene text editing by introducing FANnet, a character-level adaptive font style generation network. MC-GAN~\cite{azadi2018multi} enabled artistic glyph synthesis from only a few reference samples, generating a complete character set in a single forward pass. Subsequently, numerous studies adopted style-content disentanglement paradigms, where style and content information are encoded into separate latent representations and later combined for glyph generation. DG-Font~\cite{xie2021dg} proposed a Feature Deformation Skip Connection module that employs deformable convolutions~\cite{zhu2019deformable} to capture fine-grained geometric variations across font styles. MA-Font~\cite{qiu2024ma} further enhanced feature integration through a multi-level adaptation mechanism that progressively injects style information into content representations. FASTER~\cite{das2025faster} proposed a skeleton-mapping approach inspired by pose translation~\cite{roy2023multi}, leveraging residual downsampling and upsampling blocks to capture and propagate structural characteristics across character glyphs or words. Recent studies have increasingly leveraged CLIP~\cite{radford2021learning} embeddings to capture semantic and structural information from segmented glyph strokes. CLIP-Font~\cite{xiong2024clip} employs contrastive learning to emphasize informative regions and enhances content consistency by maximizing the cosine similarity between text-image embeddings extracted using CN-CLIP~\cite{yang2022chinese} for Chinese scripts. Moving beyond component decomposition-based methods, Patch-Font~\cite{memon2025patch} learns patch-level style representations directly from reference glyphs, enabling the synthesis of novel characters while preserving fine-grained stroke details. More recently, diffusion-based approaches have demonstrated remarkable performance in few-shot font generation. FontDiffuser~\cite{yang2024fontdiffuser} proposes a one-shot conditional diffusion framework that combines multi-scale content aggregation for preserving stroke details with style contrastive learning for consistent style transfer. To the best of our knowledge, multi-level representation, style-content disentanglement, and progressive style adaptation have largely been explored separately, while their unified modeling for multi-level style-content representation and progressive reconstruction remains less explored.

\section{Methodology} \label{sec:method}

\begin{figure}
    \centering
    \includegraphics[width=\linewidth]{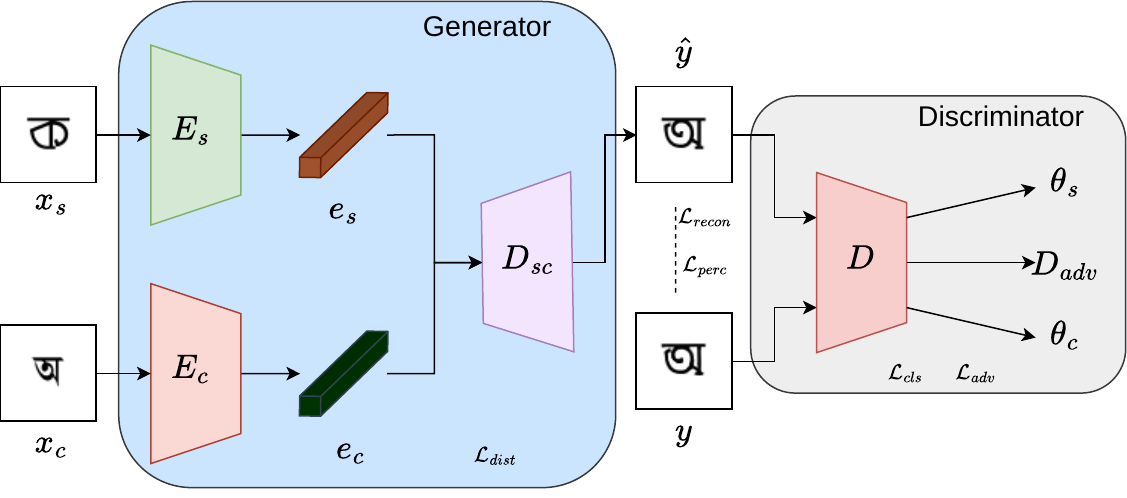}
    \caption{Overview of the proposed \textbf{CoDeFont}. The generator independently encodes the style reference $x_s$ and content reference $x_c$ using the Style Encoder $E_s$ and Content Encoder $E_c$ to obtain style embedding $e_s$ and content embedding $e_c$, respectively. The factorized representations are progressively reconciled in the Decoder $D_{sc}$ to generate the target glyph $\hat{y}$. The Discriminator provides adversarial supervision and jointly performs auxiliary style and content classification, encouraging style consistency and content preservation.}
    \label{fig:workflow}
\end{figure}

Let $\mathcal{F}$ denote a collection of $N$ font styles, where each font $f_a \in \mathcal{F}$ contains a character set $\mathcal{C}$ of $M$ characters. We denote by $x_{a,b}$ the glyph image of character $c_b \in \mathcal{C}$ rendered in font style $f_a$. Given a style reference image $x_s=x_{a,\star}\in f_a \times \mathcal{C}$ and a content reference image $x_c=x_{\star,b}\in \mathcal{F}\times c_b$, our proposed method CoDeFont generates the target glyph $\hat{y}=x_{a,b}$. Following prior FFG methods~\cite{xie2021dg, yang2024fontdiffuser, chen2025font}, the content reference is sampled from a fixed standard font $f_t$, which preserves a consistent character structure across different font styles. The Style Encoder ($E_s$) extracts a style embedding $e_s\in\mathbb{R}^{d}$ from $x_s$, while the Content Encoder ($E_c$) projects $x_c$ into a content embedding $e_c\in\mathbb{R}^{d}$. These factorized embeddings are progressively reconciled by the Decoder ($D_{sc}$) through multi-stage style conditioning to generate the target glyph $\hat{y}$, and a Discriminator ($D$) is employed to provide adversarial supervision for improving generation quality. The workflow is illustrated in Figure~\ref{fig:workflow}.

\subsection{Generator Network ($G$)}
The Generator Network consists of $E_s$, $E_c$, and $D_{sc}$, which are detailed as follows.

\subsubsection{\textbf{Style Encoder ($E_s$):}}

Given a single style reference glyph $x_s \in \mathbb{R}^{3 \times H \times W}$, stylistic cues may occur at different spatial scales, from local stroke patterns to global characteristics. The Style Encoder therefore extracts and aggregates multi-scale features into a unified style representation.

An initial convolutional block is first applied on $x_s$ to extract a low-level feature representation $f_0 \in \mathbb{R}^{C \times H \times W}$ while preserving the spatial resolution. Subsequently, $f_0$ is processed through three parallel convolutional branches followed by a residual block, with different kernel sizes, producing $f_1, f_2, f_3$ to capture stylistic cues at multiple spatial scales. These feature maps are concatenated along the channel dimension to obtain $f_4$, which is subsequently refined to produce the final fused feature representation $f_5 \in \mathbb{R}^{4C \times H \times W}$.

\begin{figure*}
    \centering
    \includegraphics[width=\linewidth]{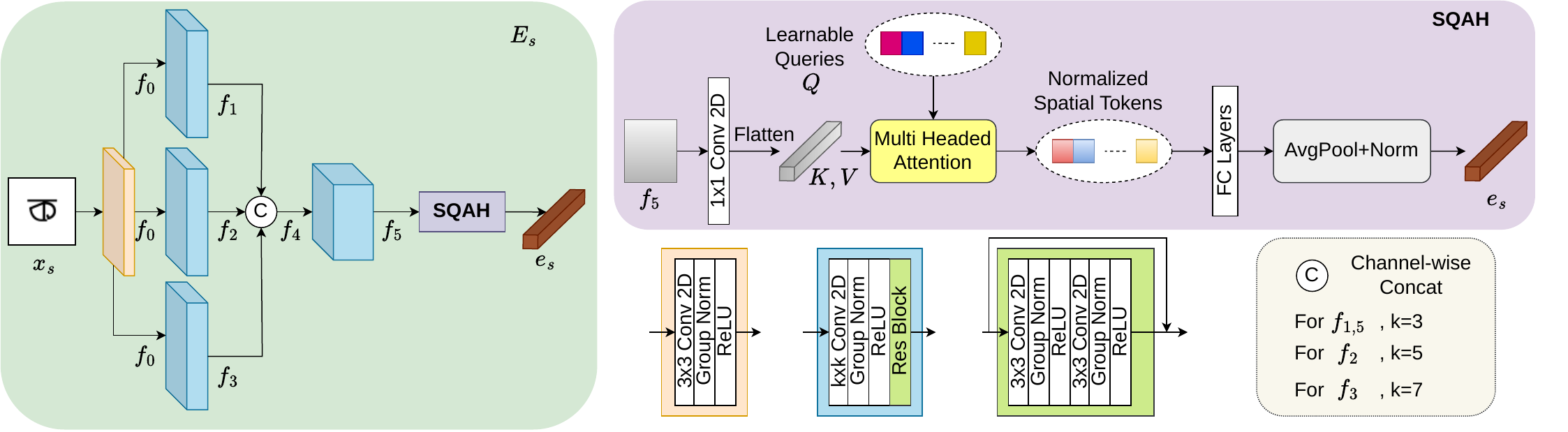}
    \caption{Architectural overview of the proposed \textbf{Style Encoder} $E_s$. Given a style image $x_s$, the encoder extracts multi-scale style features through parallel convolutional branches with different kernel sizes and fuses them into a unified representation. The fused features are then processed by the \textbf{Style-Query Attention Head (SQAH)} to obtain the final style embedding $e_s$.}
    \label{fig:styleencoder}
\end{figure*}

To selectively aggregate stylistically informative regions, we employ \textcolor{violet}{\textbf{Style-Query Attention Head (SQAH)}} a learnable query-based attention mechanism, to extract discriminative style representations from the fused feature map. Specifically, $f_5$ is projected to a $d$-dimensional embedding space and reshaped into a sequence of spatial tokens $X \in \mathbb{R}^{HW \times d}$. A set of $u$ learnable queries $Q \in \mathbb{R}^{u \times d}$ interacts with $X$ through multi-head attention to produce style tokens $T \in \mathbb{R}^{u \times d}$, where each query attends to different spatial regions. The resulting tokens are refined using Layer Normalization and fully connected layers, followed by average pooling and $\ell_2$ normalization to obtain the final style embedding $e_s$. Figure~\ref{fig:styleencoder} illustrates the architecture of $ E_s$.

\subsubsection{\textbf{Content Encoder ($E_c$):}}

Unlike style, which is inferred from the reference glyph, content must preserve the target character's structural identity independently of font appearance. Since glyph structure spans fine-grained stroke details and global spatial organization, the Content Encoder extracts representations at multiple spatial resolutions.

With the content image $x_c \in \mathbb{R}^{3 \times H \times W}$, a deformable convolutional block produces the initial feature map $g_0 \in \mathbb{R}^{2C \times \frac{H}{2} \times \frac{W}{2}}$. The deformable convolution enables adaptive sampling of local neighborhoods. Subsequently, the extracted features are progressively downsampled through three convolutional blocks, producing multi-scale feature maps $g_j \in \mathbb{R}^{2^{j+1}C \times \frac{H}{2^{j+1}} \times \frac{W}{2^{j+1}}}$, where $j \in \{1,2,3\}$. The feature maps generated through the hierarchical backbone help the encoder to capture both local stroke details and global glyph structures. To exploit multi-scale features, $\{g_1,g_2,g_3\}$ are used to capture rich structural information at different scales.

\begin{figure}
    \centering
    \includegraphics[width=\linewidth]{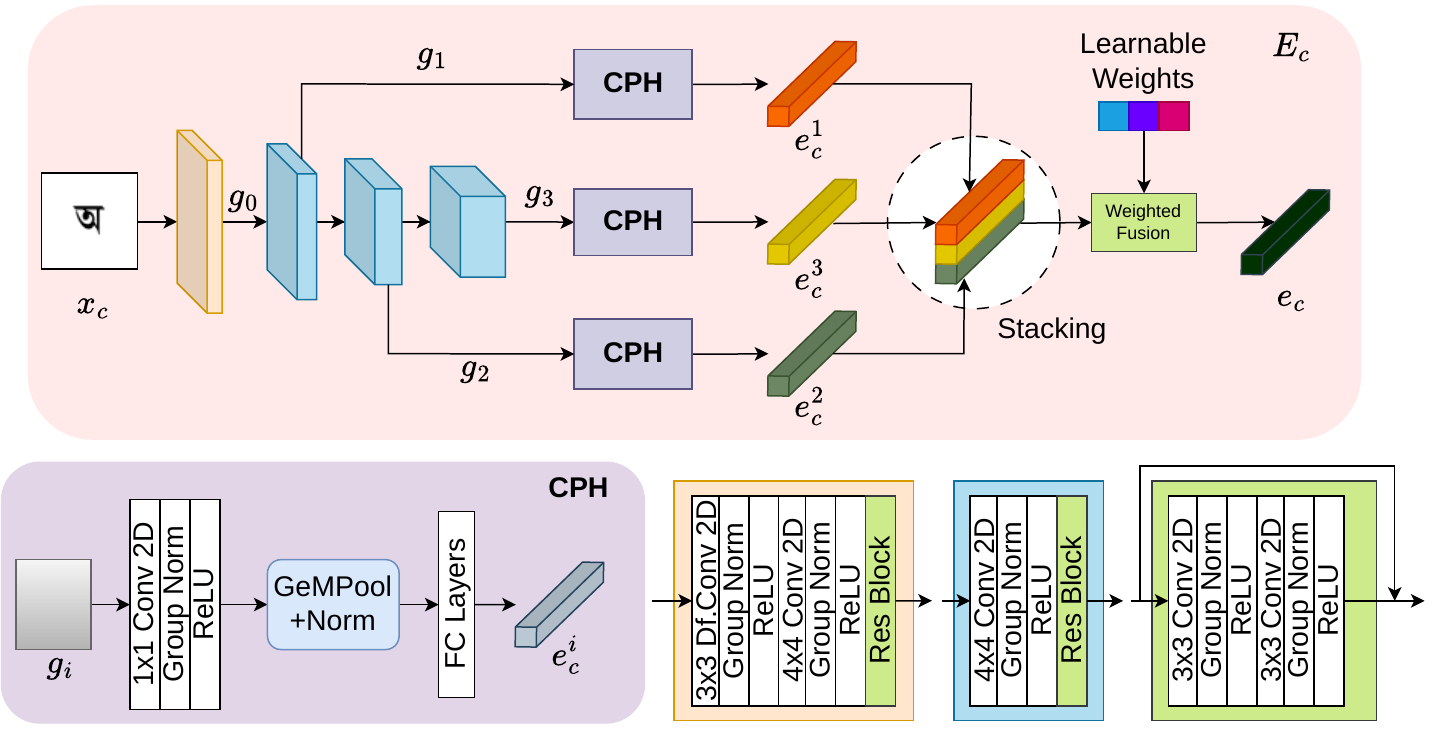}
    \caption{Architecture of the proposed \textbf{Content Encoder} $E_c$. Given a content image $x_c$, the encoder extracts multi-scale features through successive convolutional blocks. The three downsampling features are processed by the \textbf{Content Pooling Head (CPH)} and adaptively fused using learnable scale weights to obtain the final content embedding $e_c$.}
    \label{fig:contentencoder}
\end{figure}

Each feature map $g_j$ is processed by a \textcolor{violet}{\textbf{Content Pooling Head (CPH)}} consisting of a projection layer to map features into an embedding space, followed by normalization, and two fully connected layers. Finally, each scale-specific embedding is normalized to obtain $e_c^{j} \in \mathbb{R}^{d}$. Finally, the scale-specific embeddings are adaptively combined using learnable weights, allowing the network to determine the relative contribution of structural information from different resolutions., i.e., $\sum_{j=1}^{3} \alpha_j e_c^{j}$, where $\alpha_j$ denotes the learnable weight for the $j$-th embedding, with $\sum_{j=1}^{3} \alpha_j = 1$. The workflow of the proposed $E_c$ is illustrated in Figure~\ref{fig:contentencoder}.

\subsubsection{\textbf{Decoder $(D_{sc})$:}}

\begin{figure}
    \centering
    \includegraphics[width=\linewidth]{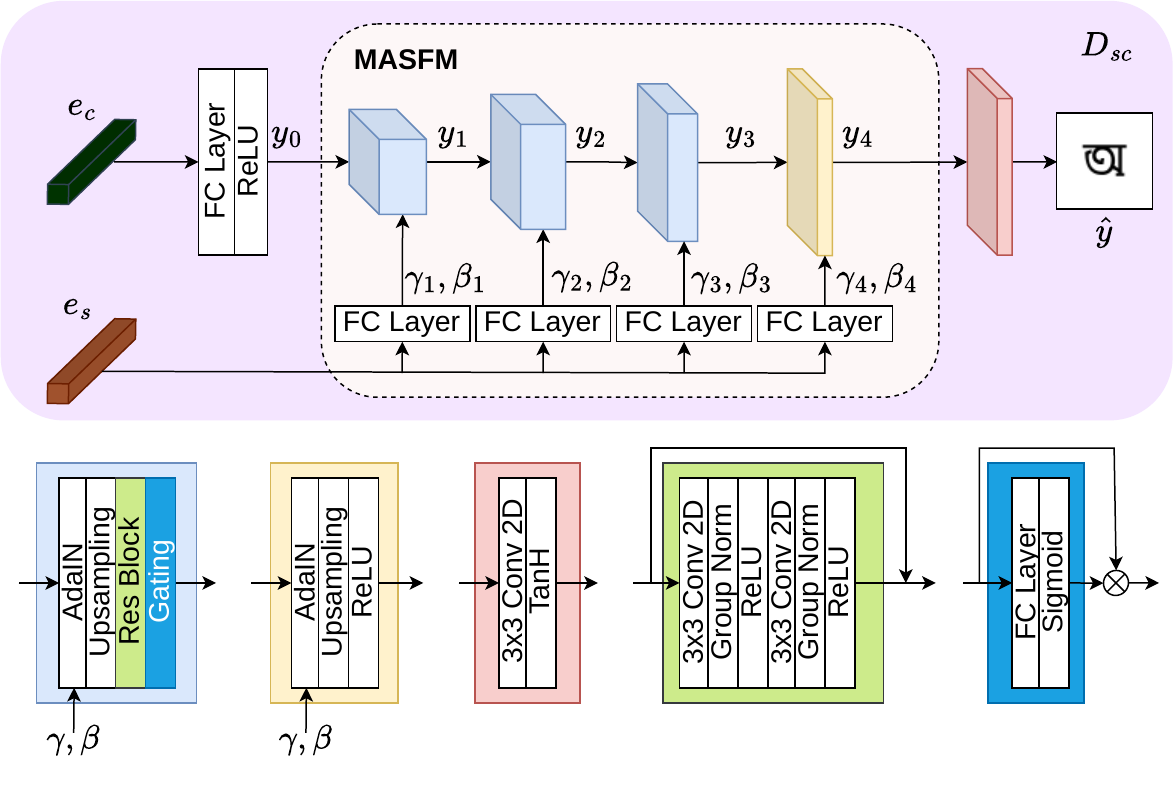}
    \caption{Architecture of the proposed \textbf{Decoder} $D_{sc}$. The decoder projects the content embedding into a low-resolution latent feature map and progressively upsamples it through the \textbf{Multi-level Adaptive Style Fusion Module (MASFM)}, which injects style information at multiple decoding levels using AdaIN~\cite{huang2017arbitrary} and channel-wise gating. The refined features are finally mapped to the target glyph $\hat{y}$ through a convolutional output layer.}
    \label{fig:decoder}
\end{figure}

The decoder combines factorized content and style representations while preserving the target glyph's structural identity. Style information is progressively introduced across decoding stages, enabling its integration with structural features at different spatial scales.

Given the content embedding $e_c$ and style embedding $e_s$, the decoder first projects $e_c$ through a fully connected layer to obtain a low-resolution latent feature map, denoted by $y_0 \in \mathbb{R}^{16C \times \frac{H}{16} \times \frac{W}{16}}$. To progressively reconcile stylistic information with the structural representation during reconstruction, the decoder employs the \textcolor{violet}{\textbf{Multi-level Adaptive Style Fusion Module (MASFM)}}, which consists of four hierarchical upsampling stages. At the $k$-th stage, where $k=\{1,2,3,4\}$, the style embedding is transformed into stage-specific affine parameters $(\gamma_k,\beta_k)$ through a linear layer. These parameters modulate the normalized input feature map $y_{k-1}$ using AdaIN~\cite{huang2017arbitrary}, thereby introducing style information at the corresponding reconstruction level. The modulated features are progressively upsampled using bilinear interpolation, convolution, and residual blocks to recover finer spatial details. After the first three stages, a channel-wise gating module uses the projected style embedding to recalibrate decoder features, complementing the statistical modulation of AdaIN, producing $y_k \in \mathbb{R}^{2^{4-k}C \times \frac{H}{2^{4-k}} \times \frac{W}{2^{4-k}}}$. Finally, the refined feature map $y_4 \in \mathbb{R}^{C \times H \times W}$ is mapped to RGB through a convolution block to generate the reconstructed glyph $\hat{y}\in\mathbb{R}^{3\times H\times W}$. The progressive decoding process is illustrated in Figure~\ref{fig:decoder}.

\subsection{Discriminator ($D$)}
We adopt the PatchGAN~\cite{isola2017image} multi-task strategy for our discriminator $D$. To produce high-quality, realistic generation, $D$ provides adversarial supervision to distinguish between fake and real images. Further, we impose auxiliary supervision~\cite{odena2017conditional} using separate classification heads for style and content. Given an input image $x' \in \mathbb{R}^{C \times H \times W}$, a convolutional feature extractor is used with spectral normalization and LeakyReLU activations is employed to learn hierarchical feature representations of size ${C_D \times H_D \times W_D}$. These shared features are then processed by an adversarial branch that produces a patch-level authenticity map, $D_{adv}(x') \in \mathbb{R}^{1 \times H_D \times W_D}$, enabling the discriminator to assess image realism across local spatial regions. In parallel, the same feature representation undergoes global average pooling before being passed to two separate prediction branches, each consisting of spectral normalization and fully connected layers, to estimate the style and content classification logits, denoted by ${\theta_{s}(x') \in \mathbb{R}^{N}, \theta_{c}(x') \in \mathbb{R}^{M}}$, respectively.

\subsection{Training Objective}
We impose a hybrid objective function with five loss components to train our proposed pipeline.

\subsubsection{\textbf{Reconstruction loss:}}
To enforce pixel-wise consistency between the generated image $\hat{y}$ and the ground-truth image $y$, we employ the $\ell_1$ reconstruction loss, defined as
\begin{equation}
\mathcal{L}_{\mathrm{recon}} =
\mathbb{E}\left[\left\| \hat{y} - y \right\|_1\right].
\end{equation}

\subsubsection{\textbf{Perceptual loss:}}

To enhance the perceptual quality of the generated image $\hat{y}$, we employ a perceptual loss~\cite{johnson2016perceptual}. Let $\phi_{\mathrm{vgg}}(\cdot) \in \mathbb{R}^{C_l \times H_l \times W_l}$ denote the feature representation extracted from the $l$-th layer of a pretrained VGG19~\cite{simonyan2014very} network. The perceptual loss is formulated as
\begin{equation}
\mathcal{L}_{\mathrm{perc}} =
\sum_{l \in \mathbb{L}=\{3,8,17,26\}}
\omega_l
\left\|
\phi_{\mathrm{vgg}}(\hat{y}) -
\phi_{\mathrm{vgg}}(y)
\right\|_1,
\end{equation}
where $\mathbb{L}$ denotes the selected VGG19 layers, and we assigned progressively lower weights $\omega_l=\{1.0,0.75,0.5,0.25\}$ for deeper layers~\cite{johnson2016perceptual}.

\subsubsection{\textbf{Adversarial loss:}}

To achieve the generation of visually realistic glyphs, we adopt a hinge-based adversarial loss. Let $D_{\mathrm{adv}}(\cdot)$ denote the discriminator output. The discriminator objective is formulated as
\begin{equation}
\begin{aligned}
\mathcal{L}_{\mathrm{adv}}^{\mathrm{D}} =
&\ \mathbb{E}_{y \sim p_{\mathrm{data}}}
\left[\max(0,\,1-D_{\mathrm{adv}}(y))\right] \\
&+\mathbb{E}_{\hat{y} \sim p_G}
\left[\max(0,\,1+D_{\mathrm{adv}}(\hat{y}))\right],
\end{aligned}
\end{equation}

while the generator is optimized using
\begin{equation}
\mathcal{L}_{\mathrm{adv}}^{\mathrm{G}}
=
-\mathbb{E}_{\hat{y} \sim p_G}
\left[D_{\mathrm{adv}}(\hat{y})\right],
\end{equation}
where $p_{\mathrm{data}}$ and $p_G$ denote the distributions of real and generated glyph images, respectively.

\subsubsection{\textbf{Auxiliary classification loss:}}

To preserve both style and content consistency in the generated glyphs, we employ an auxiliary classification loss~\cite{odena2017conditional}. Let $\theta_{s}(\cdot)$ and $\theta_{c}(\cdot)$ denote the style and content classification heads of the discriminator, respectively, and let $y_{\mathrm{label}}^{\mathrm{style}}$ and $y_{\mathrm{label}}^{\mathrm{content}}$ represent the corresponding ground-truth labels. The discriminator is optimized using
\begin{equation}
\mathcal{L}_{\mathrm{cls}}^{\mathrm{D}} =
\mathcal{L}_{\mathrm{CE}}
\left(\theta_{s}(y), y_{\mathrm{label}}^{\mathrm{style}}\right)
+
\mathcal{L}_{\mathrm{CE}}
\left(\theta_{c}(y), y_{\mathrm{label}}^{\mathrm{content}}\right),
\end{equation}

\noindent while the generator is trained with
\begin{equation}
\mathcal{L}_{\mathrm{cls}}^{\mathrm{G}} =
\mathcal{L}_{\mathrm{CE}}
\left(\theta_{s}(\hat{y}), y_{\mathrm{label}}^{\mathrm{style}}\right)
+
\mathcal{L}_{\mathrm{CE}}
\left(\theta_{c}(\hat{y}), y_{\mathrm{label}}^{\mathrm{content}}\right),
\end{equation}
where $\mathcal{L}_{\mathrm{CE}}$ denotes the cross-entropy loss.

\subsubsection{\textbf{Disentanglement loss:}}

To ensure a well-structured embedding space, a contrastive loss is imposed to encourage compact clusters for positive pairs (from the same class) and large margins between negative pairs (from different class).

For the embeddings, $e_r$, where $r \in \{s,c\}$, taking as the anchor embedding, while $e_r^{\text{pos}}$ and $e_r^{\text{neg}}$ denote the corresponding positive and negative embeddings, respectively. The cosine similarities between the positive and negative pairs are denoted by $u_{pr}$ and $u_{nr}$.

Circle Loss~\cite{sun2020circle} assigns adaptive weights $\delta_{pr}$ and $\delta_{nr}$ to positive and negative pairs, respectively, according to their optimization difficulty, with a margin hyperparameter $\eta$. During optimization, the adaptive weighting factors are treated as constants. The Circle Loss for the style embeddings is defined as
\begin{equation}
\mathcal{L}_\text{circle}^r =
\log \left(
1 +
e^{
\gamma \left(
\delta_{nr}(u_{nr}-\eta)
-
\delta_{pr}(u_{pr}-(1-\eta))
\right)
}
\right),
\end{equation}

where $\gamma$ is a scaling factor controlling the optimization strength.

Therefore, the overall disentanglement loss is given by
\begin{equation}
\mathcal{L}_\text{dist}
=
\mathcal{L}_\text{circle}^s
+
\mathcal{L}_\text{circle}^c.
\end{equation}

\subsubsection{\textbf{Total loss:}}
The overall objective follows a min–max optimization strategy between the generator $G$ and discriminator $D$, $\min_{G} \max_{D}\mathcal{L}(G, D)$, 
where the generator objective is defined as

\begin{align}
\mathcal{L}_{\mathrm{G}}
=&\;
\lambda_{\mathrm{recon}}\mathcal{L}_{\mathrm{recon}}
+\lambda_{\mathrm{perc}}\mathcal{L}_{\mathrm{perc}}
+\lambda_{\mathrm{adv}}\mathcal{L}_{\mathrm{adv}}^{\mathrm{G}} \nonumber\\
&+\lambda_{\mathrm{cls}}\mathcal{L}_{\mathrm{cls}}^{\mathrm{G}}
+\lambda_{\mathrm{dist}}\mathcal{L}_{\mathrm{dist}}.
\end{align}

\noindent
and the discriminator objective is defined as
\begin{equation}
\mathcal{L}_{\mathrm{D}} =
\lambda_{\mathrm{adv}}\mathcal{L}_{\mathrm{adv}}^{\mathrm{D}}
+ \lambda_{\mathrm{cls}}\mathcal{L}_{\mathrm{cls}}^{\mathrm{D}},
\end{equation}

\noindent
where $\lambda_\mathrm{recon}$, $\lambda_\mathrm{perc}$, $\lambda_\mathrm{adv}$, $\lambda_\mathrm{cls}$, and  $\lambda_\mathrm{dist}$ are empirically set to $5.0$, $1.0$, $0.5$, $1.0$, and $0.2$ respectively.
\section{Results} \label{sec:results}

\begin{table*}
\centering
\caption{Quantitative comparison with existing SOTA methods on Unseen and Seen Bengali and Latin fonts. The top and bottom sections correspond to Bengali and Latin, respectively. The best and second-best results are highlighted in \textbf{bold} and \underline{underlined}.}
\label{tab:quantitative_compare}
\resizebox{0.8\linewidth}{!}{
\begin{tabular}{l cccc cccc c}
\toprule

& \multicolumn{4}{c}{Unseen Fonts} 
& \multicolumn{4}{c}{Seen Fonts} & \\

\cmidrule(lr){2-5}
\cmidrule(lr){6-9}

Method 
& L1 $\downarrow$ & RMSE $\downarrow$ & SSIM $\uparrow$ & LPIPS $\downarrow$
& L1 $\downarrow$ & RMSE $\downarrow$ & SSIM $\uparrow$ & LPIPS $\downarrow$
& User Score$\uparrow$ \\

\midrule
\multicolumn{10}{c}{\textit{Bengali}} \\
\midrule

FANnet~\cite{roy2020stefann}
& 0.027 & 0.118 & 0.903 & 0.092
& 0.023 & 0.108 & 0.926 & 0.084
& 3.333 \\

MA-Font~\cite{qiu2024ma}
& 0.032 & 0.148 & 0.893 & 0.071
& 0.030 & 0.145 & 0.896 & 0.069
& 1.795 \\

PatchFont~\cite{memon2025patch}
& 0.042 & 0.181 & 0.836 & 0.057
& 0.041 & 0.178 & 0.842 & 0.053
& 3.590 \\

FontDiffuser~\cite{yang2024fontdiffuser}
& 0.020 & 0.094 & \underline{0.912} & \underline{0.047}
& \underline{0.014} & \underline{0.067} & \underline{0.954} & \underline{0.043}
& \underline{43.846} \\

FASTER~\cite{das2025faster} 
& \underline{0.019} & \textbf{0.083} & 0.881 & 0.081
& 0.018 & 0.081 & 0.887 & 0.077
& 0.769 \\

\rowcolor[HTML]{CCFFCC}
Proposed 
& \textbf{0.017} & \underline{0.090} & \textbf{0.949} & \textbf{0.030}
& \textbf{0.006} & \textbf{0.042} & \textbf{0.988} & \textbf{0.010}
& \textbf{46.667} \\

\midrule
\multicolumn{10}{c}{\textit{Latin}} \\
\midrule

FANnet~\cite{roy2020stefann}
& 0.077 & \underline{0.238} & \underline{0.731} & 0.185
& \underline{0.071} & \underline{0.229} & \underline{0.779} & \underline{0.116}
& 7.179 \\

MA-Font~\cite{qiu2024ma}
& 0.089 & 0.271 & 0.694 & 0.143
& 0.087 & 0.269 & 0.712 & 0.138
& 2.308 \\

PatchFont~\cite{memon2025patch}
& 0.098 & 0.276 & 0.675 & 0.143
& 0.094 & 0.271 & 0.681 & 0.142
& 1.795 \\

FontDiffuser~\cite{yang2024fontdiffuser}
& 0.081 & 0.267 & 0.697 & \underline{0.135}
& 0.079 & 0.261 & 0.713 & 0.127
& \underline{14.872} \\

FASTER~\cite{das2025faster} 
& \underline{0.075} & 0.246 & 0.723 & 0.139
& 0.075 & 0.239 & 0.731 & 0.137
& 8.462 \\

\rowcolor[HTML]{CCFFCC}
Proposed 
& \textbf{0.069} & \textbf{0.228} & \textbf{0.749} & \textbf{0.106}
& \textbf{0.028} & \textbf{0.128} & \textbf{0.906} & \textbf{0.044}
& \textbf{65.385} \\

\bottomrule
\end{tabular}
}
\end{table*}

\begin{figure*}
    \centering
    \includegraphics[width=0.98\linewidth]{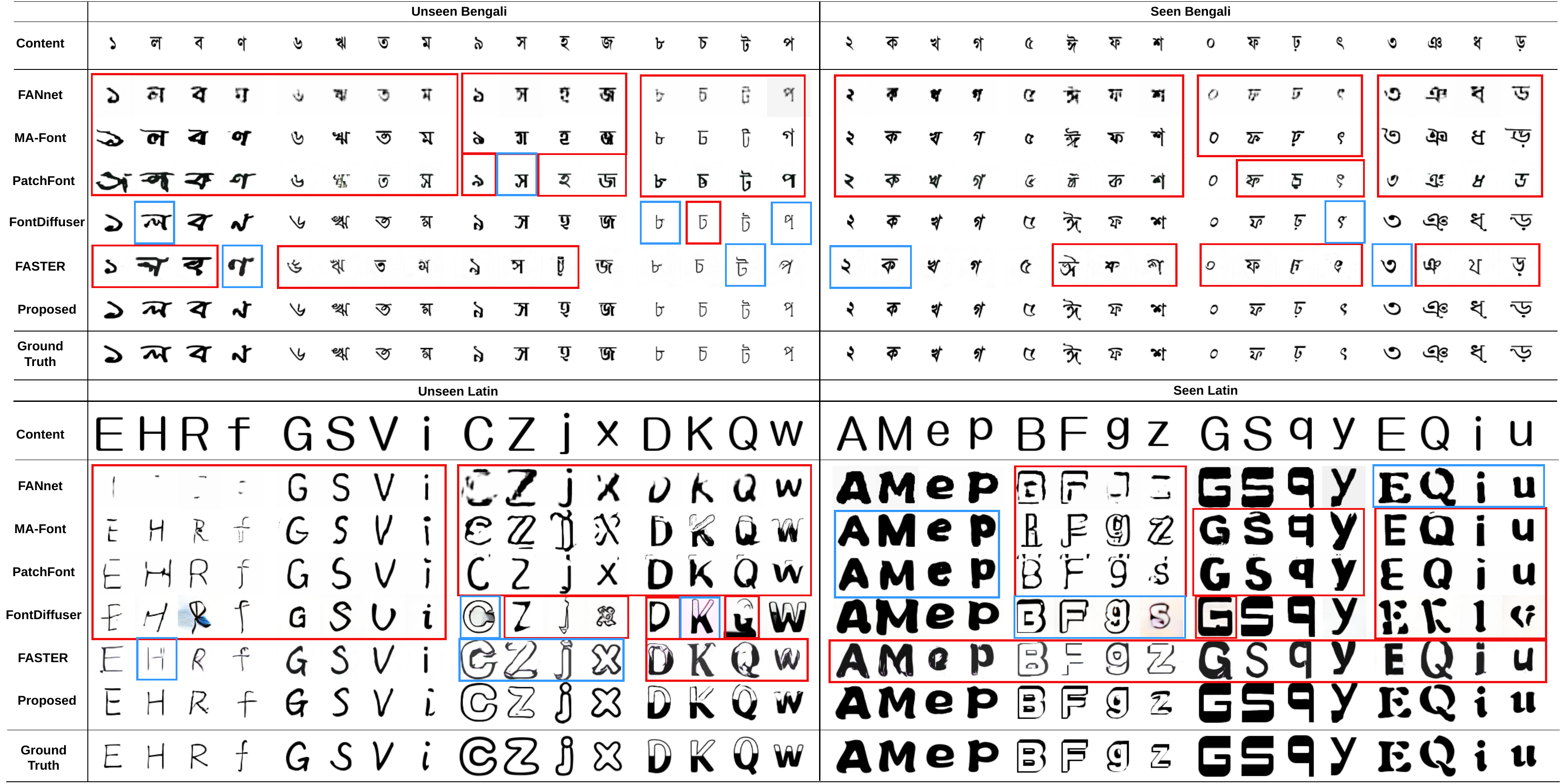}
    \caption{Qualitative comparison of our proposed method with existing SOTA techniques on Unseen and Seen Bengali and Latin fonts. The top and bottom sections correspond to Bengali and Latin, respectively. \textcolor{red}{Red} boxes highlight major failures, including poor style adaptation, structural deformation, and missing strokes, while \textcolor{blue}{blue} boxes indicate minor artifacts and structural inconsistencies for existing methods.}
    \label{fig:qualitative_compare}
\end{figure*}

\subsection{Experimental Setup}

\subsubsection{\textbf{Dataset:}}
To evaluate the efficacy of the proposed pipeline, we use two font datasets spanning structurally distinct scripts. The first is a Bengali glyph dataset comprising 595 font styles (curated and will be released), where each font contains 57 characters (10 numerals and 47 characters), with each glyph image at a spatial resolution of $64 \times 64$. The second is a Latin (English) glyph dataset~\cite{li2021few} comprising 811 unique font styles, with $52$ characters each. Both datasets are split into train, validation, and test sets with an 8:1:1 ratio.

\subsubsection{\textbf{Evaluation metrics:}}
To perform quantitative analysis of the proposed method, we employed four standard metrics as follows: \textbf{L1}, Root Mean Square Error (\textbf{RMSE}), Structural Similarity Index Measure (\textbf{SSIM})~\cite{wang2004image}, and Learned Perceptual Image Patch Similarity (\textbf{LPIPS})~\cite{zhang2018unreasonable}. L1 and RMSE assess pixel-wise differences between the generated and ground-truth images. SSIM measures perceptual similarity by considering structural information, luminance, and contrast, while LPIPS evaluates perceptual similarity in a deep feature space using feature representations extracted from a pretrained convolutional neural network.

Since no single quantitative metric can fully capture the visual quality of generated images, a user study was conducted for subjective evaluation. The study used 10 randomly selected Bangla and Latin characters (5 from each), where 78 participants were shown a ground-truth image and six corresponding images generated by six different methods, including the proposed approach. Using the ground truth as a reference, participants selected the generated image that most closely matched it. The \textbf{User Score} for each method was computed by averaging its preference rate across all participants and evaluated samples.

\subsubsection{\textbf{Implementation details:}}
We trained the proposed network for 500 epochs with a batch size of 64. For optimization, we employed the Adam optimizer with a learning rate of $2\times10^{-4}$. The style and content embeddings were represented as vectors in $\mathbb{R}^{d}$, where the embedding dimensionality was fixed at $d = 64$, and $C=64$. All experiments were conducted on a single Nvidia GeForce RTX 4080 GPU with 16GB of VRAM.

\subsection{Comparative Results}

To evaluate the efficacy of our proposed method, we compare it with several SOTA methods, including FANnet~\cite{roy2020stefann}, MA-Font~\cite{qiu2024ma}, PatchFont~\cite{memon2025patch}, FontDiffuser~\cite{yang2024fontdiffuser}, and FASTER~\cite{das2025faster}. To ensure methodological comparability, we intentionally discarded methods~\cite{xie2021dg, xiong2024clip, yao2024vq} that are solely dependent on script-specific decomposition rules.

\begin{table}[]
\centering
\caption{Comparative analysis of SOTA methods on the number of trainable parameters.}
\label{tab:parameters}
\resizebox{0.56\linewidth}{!}{
\begin{tabular}{lc}
    \toprule
    Method & \#Parameters $\downarrow$ \\
    \midrule
    FANnet~\cite{roy2020stefann} & 4.2M \\
    MA-Font~\cite{qiu2024ma} & 34.4M \\
    PatchFont~\cite{memon2025patch} & 51.3M \\
    FontDiffuser~\cite{yang2024fontdiffuser} & 323.2M \\
    FASTER~\cite{das2025faster} & 82.9M \\
    Proposed & 73.5M \\
    \bottomrule
\end{tabular}
}
\end{table}

Table~\ref{tab:quantitative_compare} shows that the proposed method achieves the best overall performance across the evaluated metrics on both Bengali and Latin scripts, with particularly consistent gains on the structurally complex Bengali script. The proposed method obtains the highest User Score on both Bengali and Latin, corresponding to an overall average preference of $56.026\%$. These results further support the visual quality and style fidelity of the generated glyphs. As illustrated in Figure~\ref{fig:qualitative_compare}, FANnet produces noisy glyphs, while MA-Font and PatchFont exhibit structural distortions, style inconsistencies, and visual artifacts. FASTER produces comparatively sharper results but still exhibits noticeable structural inaccuracies and artifacts. FontDiffuser generally generates sharper glyphs, but occasionally confuses characters with similar structures and produces noisy background regions. In contrast, the proposed method consistently generates clean glyphs with accurate character structures and faithful style characteristics across both scripts. The consistent improvements across structurally distinct scripts suggest that the proposed multi-level style-content factorization and progressive style conditioning effectively preserve character structure while transferring font-specific characteristics. Further, Table~\ref{tab:parameters} compares the number of trainable parameters of the competing methods to provide an estimate of their computational complexity.

\begin{figure*}
    \centering
    \includegraphics[width=
    0.85\linewidth]{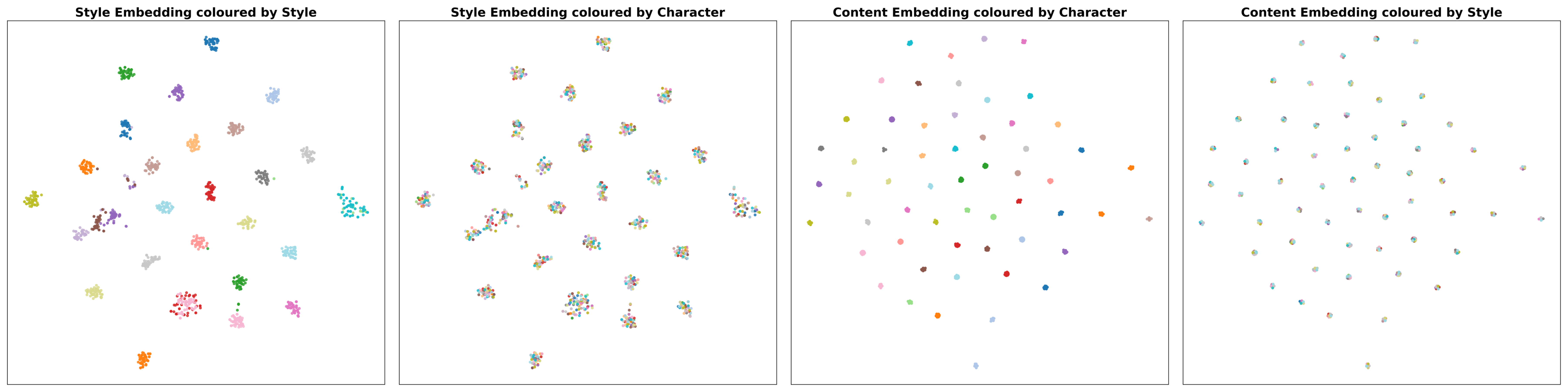}
    \caption{The t-SNE~\cite{van2008visualizing} visualization of the style ($e_s$) and content ($e_c$) embeddings, illustrating their factor-specific clustering.}
    \label{fig:tsne}
\end{figure*}

\subsection{Ablation Study}
We conduct controlled ablation studies to evaluate the proposed multi-level style-content factorization and progressive style conditioning. Unless stated otherwise, experiments use the Unseen Bengali font set.

\subsubsection{\textbf{Style-Content disentanglement:}}

To verify the effectiveness of the proposed style-content disentanglement, we visualize the embedding spaces of $e_s$ and $e_c$ using t-SNE~\cite{van2008visualizing}, as shown in Fig.~\ref{fig:tsne}. As observed, the style embeddings ($e_s$) form distinct and well-separated clusters when grouped according to style labels, while the same embeddings exhibit substantial overlap when grouped by content labels. Conversely, the content embeddings ($e_c$) produce clearly separated clusters based on content labels but become highly overlapping when visualized with style labels. 

Furthermore, to quantitatively validate the observed clustering behavior, we compute the silhouette scores~\cite{rousseeuw1987silhouettes} for all embedding-label combinations, as reported in Table~\ref{tab:tsne_silhouette}. The style embedding ($e_s$) achieves a high silhouette score when evaluated with style labels, whereas its score becomes negative when evaluated with content labels. Conversely, the content embedding ($e_c$) attains an almost perfect silhouette score with content labels while yielding a near-zero negative score with style labels. 
\begin{table}[!htbp]
    \centering
    \caption{Silhouette scores of the style and content embeddings under different label assignments}
    \label{tab:tsne_silhouette}
    \resizebox{0.8\linewidth}{!}{
    \begin{tabular}{cccc}
        \toprule
        $e_s$ by Style & $e_s$ by Content & $e_c$ by Style & $e_c$ by Content \\
        \midrule
        0.695 & -0.055 & -0.017 & 0.999 \\
        \bottomrule
    \end{tabular}
    }
\end{table}

These results show that $e_s$ is primarily associated with style and largely insensitive to content, while $e_c$ exhibits the complementary behavior, supporting the proposed factor-specific representation learning strategy.

\subsubsection{\textbf{Embedding dimension:}}
To determine an appropriate embedding dimension, we evaluate the proposed method with different embedding sizes. As shown in Table~\ref{tab:dimension}, the best performance is achieved at $d=64$. Smaller embeddings suffer from limited representational capacity, while larger ones degrade performance due to increased model complexity and overfitting.

\begin{table}[!htbp]
    \centering
    \caption{Effect of embedding dimension on the performance of our proposed method. Best results are shown in \textbf{bold}.}
    \label{tab:dimension}
    \resizebox{0.8\linewidth}{!}{
    \begin{tabular}{ccccc}
        \toprule
        Emb. Dim. & L1 $\downarrow$ & RMSE $\downarrow$ & SSIM $\uparrow$ & LPIPS $\downarrow$ \\
        \midrule
         32  & 0.018 & 0.095 & 0.945 & 0.033 \\
        \textbf{64}  & \textbf{0.017} & \textbf{0.090} & \textbf{0.949} & \textbf{0.030} \\
        128 & \textbf{0.017} & 0.091 & 0.948 & 0.031 \\
        256 & 0.018 & 0.096 & 0.944 & 0.034 \\
        512 & 0.033 & 0.151 & 0.886 & 0.077 \\
        \bottomrule
    \end{tabular}
    }
\end{table}

\subsubsection{\textbf{Impact of loss components:}}
To evaluate the proposed loss functions, we conduct an ablation study by training the model with and without $\mathcal{L}_\text{cls}$ and $\mathcal{L}_\text{dist}$ while keeping the remaining losses unchanged. As shown in Table~\ref{tab:loss_ablation}, removing either objective degrades generation quality, while jointly incorporating $L_{cls}$ and $L_{dist}$ yields the best performance. This indicates that explicit supervision of factor identity and embedding separation provides complementary benefits to image-level reconstruction and perceptual objectives.

\begin{table}[h]
\centering
\caption{Quantitative analysis of the effectiveness of $\mathcal{L}_{\text{cls}}$ and $\mathcal{L}_{\text{dist}}$ on the generation quality. Better results are shown in \textbf{bold}.}
\resizebox{0.8\linewidth}{!}{
\begin{tabular}{cc cccc}
\toprule
\multicolumn{2}{c}{Loss Function} & \multicolumn{4}{c}{Metrics} \\
\cmidrule(lr){1-2} \cmidrule(lr){3-6}
$\mathcal{L}_{\text{dist}}$ & $\mathcal{L}_{\text{cls}}$
& L1 $\downarrow$
& RMSE $\downarrow$
& SSIM $\uparrow$
& LPIPS $\downarrow$ \\
\midrule
\xmark & \xmark & 0.034 & 0.251 & 0.852 & 0.153 \\
\xmark & \cmark & 0.029 & 0.219 & 0.874 & 0.132 \\
\cmark & \xmark & 0.022 & 0.168 & 0.885 & 0.072 \\
\cmark & \cmark & \textbf{0.017} & \textbf{0.090} & \textbf{0.949} & \textbf{0.030} \\
\bottomrule
\end{tabular}
}
\label{tab:loss_ablation}
\end{table}

\subsubsection{\textbf{Impact of style gating:}}

To evaluate the proposed style-gating mechanism, we compare the full decoder with a variant without style gating while keeping all other settings unchanged. As shown in Table~\ref{tab:gating}, the style-gating module consistently improves performance across all evaluation metrics, demonstrating that channel-wise feature modulation enhances style transfer while preserving the underlying glyph structure.

\begin{table}[h]
    \centering
    \caption{Effect of gating mechanism on the performance of our proposed method. Better results are shown in \textbf{bold}.}
    \label{tab:gating}
    \resizebox{0.8\linewidth}{!}{
    \begin{tabular}{ccccc}
        \toprule
        Gating & L1 $\downarrow$ & RMSE $\downarrow$ & SSIM $\uparrow$ & LPIPS $\downarrow$ \\
        \midrule
         \xmark  & 0.019 & 0.097 & 0.938 & 0.038 \\
        \cmark  & \textbf{0.017} & \textbf{0.090} & \textbf{0.949} & \textbf{0.030} \\
        \bottomrule
    \end{tabular}
    }
\end{table}

\subsubsection{Impact of multi-scale feature:}

As shown in Table~\ref{tab:ms_ablation}, multi-scale representations benefit both style and content modeling. Removing them from both encoders yields the poorest performance, while enabling them in either encoder improves the results. Using multi-scale representations in both encoders achieves the best performance across all metrics, highlighting the benefit of jointly modeling complementary multi-level style and structural information.

\begin{table}[h]
\centering
\caption{Quantitative analysis of the effectiveness of multi-scale (MS) features in the style encoders $E_s$ and $E_c$ on generation quality. Best results are shown in \textbf{bold}.}
\resizebox{0.8\linewidth}{!}{
\begin{tabular}{cc cccc}
\toprule
\multicolumn{2}{c}{MS Feature} & \multicolumn{4}{c}{Metrics} \\
\cmidrule(lr){1-2} \cmidrule(lr){3-6}
$E_s$ & $E_c$
& L1 $\downarrow$
& RMSE $\downarrow$
& SSIM $\uparrow$
& LPIPS $\downarrow$ \\
\midrule
\xmark & \xmark & 0.029 & 0.192 & 0.932 & 0.121 \\
\xmark & \cmark & 0.021 & 0.154 & 0.939 & 0.075 \\
\cmark & \xmark & 0.019 & 0.108 & 0.943 & 0.054 \\
\cmark & \cmark & \textbf{0.017} & \textbf{0.090} & \textbf{0.949} & \textbf{0.030} \\
\bottomrule
\end{tabular}
}
\label{tab:ms_ablation}
\end{table}
\section{Conclusion}\label{sec:conclusion}

In this paper, we introduced \textbf{CoDeFont}, a one-shot font generation framework that performs multi-level style-content factorization and progressively reconciles style information during decoding. Extensive experiments demonstrate that CoDeFont consistently outperforms existing state-of-the-art methods for Bengali font generation while generalizing effectively to Latin scripts. Comprehensive ablation studies further validate the effectiveness of disentangled embedding spaces, multi-level representations, factor-specific learning, and progressive style conditioning. In future work, we plan to extend CoDeFont to more challenging scenarios, including cross-script and multilingual font generation.

{
    \small
    \bibliographystyle{ieeenat_fullname}
    \bibliography{main}
}


\end{document}